\documentclass[letterpaper, 10 pt, conference]{ieeeconf}  

\IEEEoverridecommandlockouts                              

\usepackage{amsmath,graphicx,amssymb,epsfig,mathrsfs,gensymb}
\usepackage{hyperref,verbatim}

\title{\LARGE \bf
3D-Aware Object Localization using Gaussian Implicit Occupancy Function
}

\author{Vincent Gaudilli\`ere \quad Leo Pauly \quad Arunkumar Rathinam \quad Albert Garcia Sanchez \\ Mohamed Adel Musallam \quad Djamila Aouada
\thanks{This work was partly funded by the Luxembourg National Research Fund (FNR) under the project reference BRIDGES2020/IS/14755859/MEET-A/Aouada.}
\thanks{The authors are with SnT - Interdisciplinary Center for Security, Reliability and Trust, University of Luxembourg. Contact email: {\tt\small vincent.gaudilliere@uni.lu}}%
}
\usepackage{floatrow}
\usepackage[normalem]{ulem}

\begin{document}

\maketitle
\thispagestyle{empty}
\pagestyle{empty}

\begin{abstract}


To automatically localize a target object in an image is crucial for many computer vision applications. To represent the 2D object, ellipse labels have recently been identified as a promising alternative to axis-aligned bounding boxes. This paper further considers 3D-aware ellipse labels, \textit{i.e.}, ellipses which are projections of a 3D ellipsoidal approximation of the object, for 2D target localization. Indeed, projected ellipses carry more geometric information about the object geometry and pose (3D awareness) than traditional 3D-agnostic bounding box labels. Moreover, such a generic 3D ellipsoidal model allows for approximating known to coarsely known targets. We then propose to have a new look at ellipse regression and replace the discontinuous geometric ellipse parameters with the parameters of an implicit Gaussian distribution encoding object occupancy in the image. The models are trained to regress the values of this bivariate Gaussian distribution over the image pixels using a statistical loss function. We introduce a novel non-trainable differentiable layer, E-DSNT, to extract the distribution parameters. Also, we describe how to readily generate consistent 3D-aware Gaussian occupancy parameters using only coarse dimensions of the target and relative pose labels. We extend three existing spacecraft pose estimation datasets with 3D-aware Gaussian occupancy labels to validate our hypothesis. Labels and source code are publicly accessible here: \href{https://cvi2.uni.lu/3d-aware-obj-loc/}{https://cvi2.uni.lu/3d-aware-obj-loc/}.

\end{abstract}

\section{INTRODUCTION} 
\label{sec:intro}




 Object localization in images has gained interest within the computer vision community due to its potential impact on a wide range of applications. While the axis-aligned bounding box has been the de facto standard representation for object detections~\cite{liu2020deep}, ellipses have been recently identified as another generic representation able to carry more information about the object projection, such as its orientation and more fitted envelope, therefore enabling, for instance, more accurate 3D reconstructions~\cite{DongRPI21}.

\begin{figure}[t]
    \centering
    \includegraphics[trim=11.2cm 1.5cm 7cm 0.7cm,clip,width=.85\linewidth]{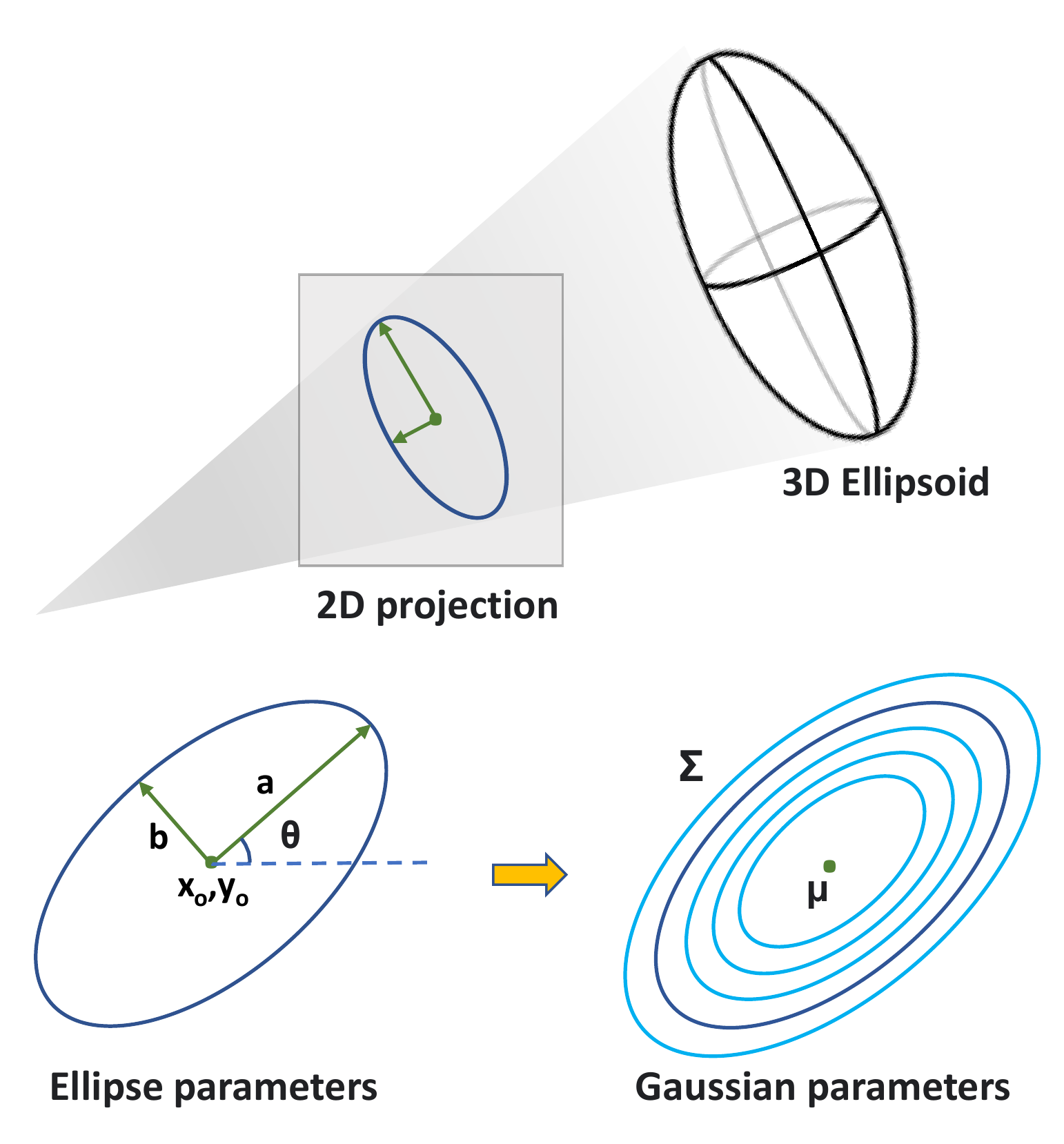}
    \caption{\textbf{Top:} We consider 2D projections of a 3D ellipsoid approximating the object. \textbf{Bottom-left:} Standard ellipse representation with centre coordinates ($x_0,y_0$), major ($a$) and minor ($b$) axes, and orientation $\theta$. \textbf{Bottom-right:} To cope with the discontinuity of $\theta$, we propose to represent the ellipse by an underlying implicit Gaussian distribution parameterized by both continuous mean $\mu=\left(x_0,y_0\right)$ and covariance $\Sigma=\begin{pmatrix}
        \mathrm{cos}(\theta) & -\mathrm{sin}(\theta) \\
        \mathrm{sin}(\theta) & \mathrm{cos}(\theta)
    \end{pmatrix}\begin{pmatrix}
        a^2 & 0 \\
        0 & b^2
    \end{pmatrix}\begin{pmatrix}
        \mathrm{cos}(\theta) & \mathrm{sin}(\theta) \\
        -\mathrm{sin}(\theta) & \mathrm{cos}(\theta)
    \end{pmatrix}$.}
    \label{fig:ellipse-params}
\end{figure}

\textbf{2D Ellipse Regression} Pioneering work in ellipse regression has been focusing on geometric ellipse parameters, \textit{i.e.} centre coordinates, minor and major axes and orientation angle (see Fig. \ref{fig:ellipse-params}). To regress these parameters, Ellipse Proposal Networks~\cite{WangDR0XX19} use L1 losses, thus requiring relative weighting between location and axes errors on one side and orientation error on the other. To circumvent this, Lin \textit{et al.}~\cite{LinLSSS19} transform the parameters so that they are in the same order of magnitude. The main limitation of these methods lies in the discontinuity of the angular value to regress.

Gaussian Proposal Networks (GPN)~\cite{Li19}, inspired by Region Proposal Networks (RPN)~\cite{RenHGS15}, look at ellipses as 2D Gaussian distributions on the image plane and minimizes the Kullback-Leibler (KL) divergence between the proposed and groundtruth distributions as one single loss. Since KL divergence has an analytical form for Gaussians and is differentiable, GPN can be easily implemented and trained with a back-propagation algorithm. However, the Gaussian parameters are derived from the regressed geometric parameters (position, axes and orientation), therefore facing the same discontinuity issue on angle regression. GPN has been integrated into an object detection pipeline by Pan \textit{et al.}~\cite{PanFWZR21}, where the Wasserstein metric is used instead of KL to provide the model with proper distance loss. Ellipse R-CNN~\cite{DongRPI21} also regresses geometric ellipse parameters with enhanced occlusion robustness, while ElDet~\cite{Wang_2022_ACCV} adds training objectives such as maximizing Intersection-over-Union (IoU) score between predicted and groundtruth ellipses. However, these methods are designed to detect naturally occurring elliptic shapes in images, which does not correspond to our use-case, where ellipses originate from a 3D virtual ellipsoid modeling the object.

\textbf{3D-Aware Ellipse Regression} To the best of our knowledge, Zins \textit{et al.}~\cite{ZinsSB20,ZinsSB22_IJCV,ZinsSB22_IROS} were the first ones to regress 3D-aware ellipses. In \cite{ZinsSB20}, they apply a L2 loss on the ellipse centre and dimensions, while the angle prediction is framed as a classification problem with posterior angular correction loss. They also proposed two types of implicit functions characterized by ellipse parameters for robustifying the loss function: \textit{a local signed distance function} enabling pixel-to-pixel comparison between groundtruth and predicted functions values~\cite{ZinsSB22_IJCV} and \textit{an algebraic distance function} based on ellipse equation combined with an adaptive sampling to provide rotation invariance~\cite{ZinsSB22_IROS}. However, implicit function values are still computed based on the regressed geometric ellipse parameters.

In this paper, we take up this implicit function idea but propose to continuously regress its values over the image pixels in the form of an occupancy heatmap. Before training, the Gaussian distribution parameters and heatmaps are directly computed from the ellipsoid projections using the relative object-camera poses. We extract these parameters from the regressed heatmap during forward propagation thanks to a novel non-trainable differentiable layer: \textit{Extended-Differentiable Spatial to Numerical Transform} (E-DSNT). A combination of statistical losses is finally proposed to optimize the model. Though our method can be used in any use-case requiring 2D target localization, we focus on the Space Situational Awareness (SSA) application to validate it in this paper. Indeed, automatically localizing a target uncooperative spacecraft is crucial for tasks such as in-orbit rendezvous. We evaluated our work on three spacecraft pose estimation benchmark datasets. In a nutshell, our contributions are three-fold:
\begin{itemize}
    \item A novel and fully differentiable object localization pipeline that can regress 3D-aware ellipse labels directly from an image. This proposed approach achieves state-of-the-art performance on standard spacecraft localization benchmarks;
    \item A method for generating 3D-aware Gaussian occupancy labels given only 6-Degree-of-Freedom relative poses and coarse object dimensions;
    \item An open-access release of 3D-aware Gaussian occupancy labels (heatmaps, mean and covariance labels) for three existing spacecraft pose estimation datasets.
\end{itemize}

The organization of the rest of the paper is as follows. Section \ref{sec:labels} describes the generation of 3D-aware Gaussian occupancy labels for object localization in images. Section \ref{sec:model} presents our object Localization model designed to regress such labels. Then, experimental comparisons demonstrating the state-of-the-art performance of the method are provided in Section \ref{sec:xp}. Finally, Section \ref{sec:concl} concludes the paper.



\section{3D-AWARE GAUSSIAN OCCUPANCY LABELS}
\label{sec:labels}

\begin{figure}[t]
    \centering
    \includegraphics[width=\linewidth]{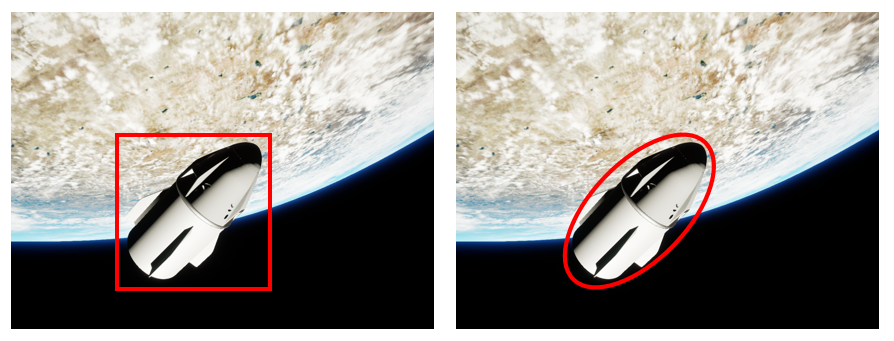}
    \caption{Illustration of bounding box object label (left) and ellipse label (right). The ellipse is a more accurate representation of many man-based objects. Image from the URSO Dataset~\cite{ProencaG20}.}
    \label{fig:bbox-vs-ellipse}
\end{figure}

\begin{figure*}[t]
    \centering
    \includegraphics[trim=0cm 0.2cm 0cm 0cm, clip, width=\linewidth]{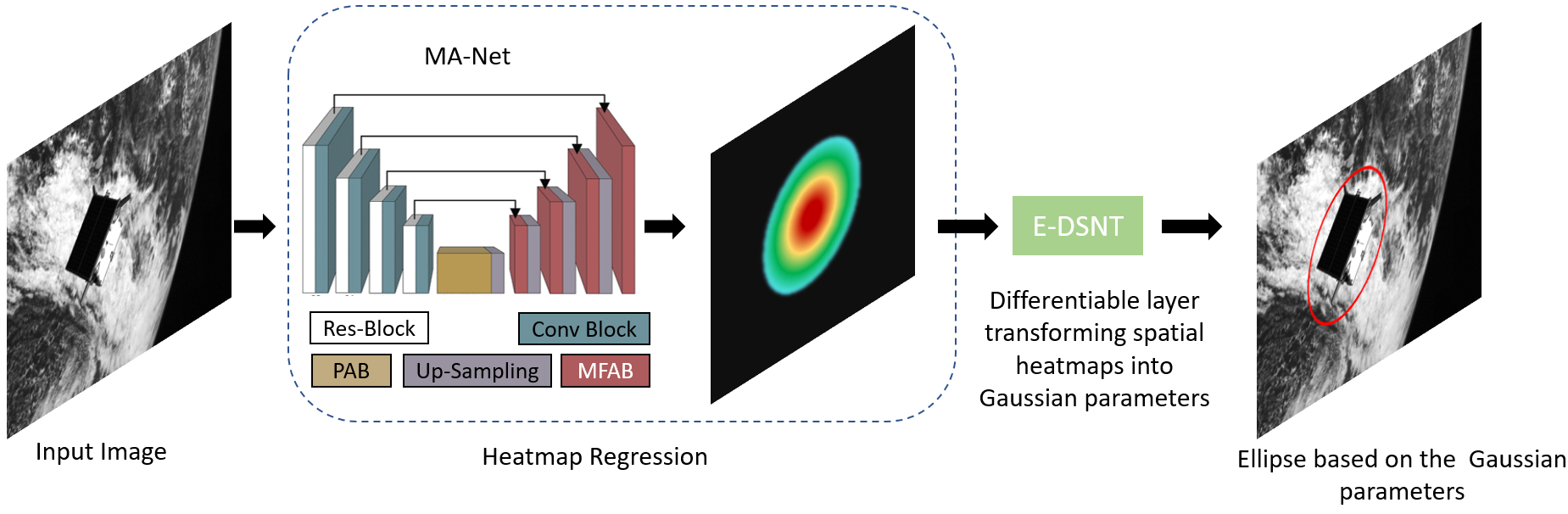}
    \caption{Illustration of our object localization pipeline designed to regress the parameters of a 3D-aware Gaussian implicit occupancy function (represented as a heatmap in the central picture) in a fully differentiable manner. The model combines an MA-Net backbone to regress the implicit function values over image pixels with a novel non-trainable differentiable layer (E-DSNT) to extract Gaussian distribution parameters. The corresponding ellipse is then derived (overlaid in red in the right picture). Image from the SPEED+ Dataset~\cite{SPEED+}.}
    \label{fig:model_large}
\end{figure*}

This section describes how to readily generate Gaussian occupancy labels from a 3D ellipsoidal model of the object and the relative camera-object pose.

\subsection{2D Object Occupancy}

Object occupancy in a picture can be defined as the set of pixels corresponding to that object. In most cases, object detection labels $L_{obj}$ are in the form of bounding boxes designed to encompass the object occupancy (see Fig. \ref{fig:bbox-vs-ellipse}). Considering a bounding box with centre $(x_c, y_c)$ and dimensions $(W, H)$, object detection labels can be written:
\begin{equation}
    L_{obj}^{(bbox)} = \left[x_c, y_c, W, H\right]\mbox{.}
\end{equation}

Now assuming that the object occupancy is encoded by an implicit occupancy function $O_{\mathbf{p}}:\mathbb{R}^2\rightarrow\mathbb{R}$ characterized by parameters $\mathbf{p}$, object labels can be written
\begin{equation}
    L_{obj} = \mathbf{p}\mbox{.}
\end{equation}

\subsection{2D Gaussian Occupancy Labels}

Statistical information about the object occupancy is given by means $\mu_x,\mu_y$, variances $\Sigma_{xx},\Sigma_{yy}$ and covariance $\Sigma_{xy}$ of the occupied pixels in the 2D image. Therefore, a natural implicit occupancy function is the bivariate Gaussian distribution with centre $\mathbf{\mu}=(\mu_x,\mu_y)^\top$ and covariance matrix $\Sigma=\begin{pmatrix}
    \Sigma_{xx} & \Sigma_{xy} \\
    \Sigma_{xy} & \Sigma_{yy}
\end{pmatrix}\mbox{.}$  Finally, we define Gaussian occupancy labels as
\begin{equation}
\label{eq:L_obj_Gauss}
    L_{obj}^{(Gauss)} = [\mu_x, \mu_y, \Sigma_{xx}, \Sigma_{yy}, \Sigma_{xy}]\mbox{.}
\end{equation}

\subsection{3D-Aware Gaussian Occupancy Labels}

The parameters of a bivariate Gaussian distribution are those of a characteristic ellipse, and vice versa. In our method, we are interested in regressing the parameters of the elliptic projection of a 3D ellipsoid, and for this reason, we consider the Gaussian distribution arising from it.

More precisely, given an object whose dimensions along its three orthogonal principal directions are $2a', 2b', 2c'$, an ellipsoidal approximation of the object is characterized by matrix
\begin{equation}
    Q^{*}= diag\begin{pmatrix} 
    a'^2 & b'^2 & c'^2 & -1 
\end{pmatrix}\mbox{.}
\end{equation}

Then, denoting $\mathrm{T}$ the object centre position and $\mathrm{R}$ the object orientation with respect to the camera, the ellipsoid projection into the image (ellipse) is given by matrix
\begin{equation}
    C^{*}=PQ^{*}P^\top\mbox{,}
\end{equation}
where $P=\mathrm{K}\left[\mathrm{R}|\mathrm{T}\right]$ and $\mathrm{K}$ the camera intrinsic matrix~\cite{Hartley2004}.




The projected ellipse $C^{*}$ is therefore in the form:
\begin{equation}
    C^{*}=s\begin{pmatrix}
        \cdot & \cdot & -x_0 \\
        \cdot & \cdot & -y_0 \\
        -x_0 & -y_0 & -1
    \end{pmatrix}\mbox{,}
\end{equation}
where $s$ is a scale factor, $\cdot$ denotes any value, and $(x_0,y_0)$ is the ellipse centre that corresponds to the peak location of the underlying Gaussian occupancy function:
\begin{equation}
    \mu=(\mu_x,\mu_y)^\top=(x_0,y_0)^\top\mbox{.}
\end{equation}
The centred ellipse is then obtained by
\begin{equation}
    C_\mathrm{centered}^{*}=TC^{*}T^\top\mbox{,}
\end{equation}
where $
    T=\begin{pmatrix}
        1 & 0 & -x_0 \\
        0 & 1 & -y_0 \\
        0 & 0 & 1
    \end{pmatrix}\mbox{.} $ \\
    
The $2\times2$ upper-left part of $C_\mathrm{centered}^{*}$ corresponds to the covariance $\Sigma$ of the Gaussian occupancy function, and its eigendecomposition provides the ellipse orientation $\theta$ and semi-axes $(a,b)$:
\begin{align}
    \Sigma&=\begin{pmatrix}
    \Sigma_{xx} & \Sigma_{xy} \\
    \Sigma_{xy} & \Sigma_{yy}
\end{pmatrix}\\
    &=\begin{pmatrix}
        \mathrm{cos}(\theta) & -\mathrm{sin}(\theta) \\
        \mathrm{sin}(\theta) & \mathrm{cos}(\theta)
    \end{pmatrix}\begin{pmatrix}
        a^2 & 0 \\
        0 & b^2
    \end{pmatrix}\begin{pmatrix}
        \mathrm{cos}(\theta) & \mathrm{sin}(\theta) \\
        -\mathrm{sin}(\theta) & \mathrm{cos}(\theta)
    \end{pmatrix}\mbox{.}\label{eq:eig}
\end{align}


\section{OBJECT LOCALIZATION}
\label{sec:model}

Our object localization pipeline, illustrated in Fig.\ref{fig:model_large}, can regress Gaussian occupancy labels $L_{obj}^{(Gauss)}$ via implicit function estimation in a fully differentiable manner. We introduce a differentiable heatmap parameters extraction layer E-DSNT coupled with a heatmap regression network to estimate the implicit function values. These modules are respectively presented in Sections \ref{ssec:E-DSNT} and \ref{ssec:hm-reg}. The statistical loss used to optimize the model is presented in Section \ref{ssec:losses}.
\subsection{Heatmap Regression Network}
\label{ssec:hm-reg}

Any state-of-the-art segmentation network can be used to regress the Gaussian occupancy function values $O_{\mathbf{p}}(x,y)$ across image pixels. In the experiments, we use the MA-Net~\cite{FanWLW20} network with ResNet34~\cite{He_2016_CVPR} backbone since this model can capture contextual dependencies based on an attention mechanism, using two blocks (refer to Fig.\ref{fig:model_large}): a Position-wise Attention Block (PAB), which captures the spatial dependencies between pixels in a global view, and a Multi-scale Fusion Attention Block (MFAB), which captures the channel dependencies between any feature map by multi-scale semantic feature fusion. As a last layer, a flattening-softmax layer is added to ensure a normalized probability density function, which we refer to as a \textit{heatmap} in what follows.

\subsection{Differentiable Extraction of Gaussian Occupancy Labels}
\label{ssec:E-DSNT}



A bivariate Gaussian distribution is characterized by its mean value $\mu$, being the coordinates of the peak, and its covariance matrix $\Sigma$ encoding the spatial extent of the distribution. A differentiable mean extraction layer was previously introduced in the literature (see Section \ref{sssec:DSNT}), and we extend it to extract the additional covariance matrix values, refer to Section \ref{sssec:cov-ext}.

\vspace{0.1cm}
\subsubsection{Mean Extraction}
\label{sssec:DSNT}

Differentiable Spatial to Numerical Transform (DSNT)~\cite{DSNT} is a non-trainable and differentiable
layer for extracting the mean value of a given normalized Gaussian heatmap $\hat{Z}$ with size $H \times W$. Following this method, we compute two coordinates encoding matrices $X$ and $Y$ with entries
\begin{equation*}
\begin{split}
    X_{i,j} = \frac{2j-(W+1)}{W}\mbox{;} \hspace{0.2cm} Y_{i,j} = \frac{2i-(H+1)}{H}\mbox{,} \\
    \forall i = 1,\ \dots,\ H\mbox{;} \hspace{0.2cm} \forall j = 1,\ \dots,\ W\mbox{.}
\end{split}
\end{equation*}

Observing that the heatmap encodes the probability $P$ of the pixel $\left[X_{i,j}\mbox{ , }Y_{i,j}\right]$ to be the location of the peak $p$, we have:
\begin{equation*}
\begin{split}
    P(p=[X_{i,j}\mbox{ , }Y_{i,j}]) = \hat{Z}_{i,j}\mbox{.}
\end{split}
\end{equation*}

The prediction of $p$ is made through its expectation
\begin{equation}
\label{eq:mu}
\begin{split}
    \mu = \mathbb{E}[p] = \left[<\hat{Z}, X>_F\quad <\hat{Z}, Y>_F \right]
\end{split}
\end{equation}
where $<\cdot,\cdot>_F$ is the Frobenius inner product.

\vspace{0.1cm}
\subsubsection{Covariance Matrix Extraction}
\label{sssec:cov-ext}


Denoting $p_x,p_y$ the coordinates of $p$, we extend DSNT to extract the heatmap variances along the x-axis $\Sigma_{xx}$ and y-axis $\Sigma_{yy}$, as well as its covariance value $\Sigma_{xy}$. We refer to such parameter extraction procedure as \textit{Extended}-DSNT (E-DSNT). Specifically, we use the definitions of the aforementioned quantities and derive the following equations: 
\begin{equation}
\label{eq:sigma-xx}
\begin{split}
    \Sigma_{xx} &= \mathbb{E}[(p_x-\mathbb{E}[p_x])^2] \\&= <\hat{Z}, (X-\mu_x) \odot (X-\mu_x)>_F
\end{split}
\end{equation}
\begin{equation}
\label{eq:sigma-yy}
\begin{split}
    \Sigma_{yy} &= \mathbb{E}[(p_y-\mathbb{E}[p_y])^2] \\&= <\hat{Z}, (Y-\mu_y) \odot (Y-\mu_y)>_F
\end{split}
\end{equation}
\begin{equation}
\label{eq:sigma-xy}
\begin{split}
    \Sigma_{xy} &= \mathbb{E}[(p_x-\mathbb{E}[p_x])(p_y-\mathbb{E}[p_y])] \\&= <\hat{Z}, (X-\mu_x) \odot (Y-\mu_y)>_F
\end{split}
\end{equation}

Therefore, the Gaussian distribution parameters are the centre $\mu$, obtained from Eq. \eqref{eq:mu}, and the covariance matrix $\Sigma$ formed by left terms of Eq. \eqref{eq:sigma-xx}, \eqref{eq:sigma-yy} and \eqref{eq:sigma-xy}. 

\vspace{0.1cm}
\subsubsection{Ellipse Parameters Computation}
The centre of the ellipse is simply $\left(x_0,y_0\right)=\mu$. Its axes $a,b$ and orientation $\theta$ are obtained by the eigendecomposition of the covariance matrix $\Sigma$ (see Eq. \ref{eq:eig}).

\subsection{Model Loss}
\label{ssec:losses}

Our loss $\mathcal{L}$ can be written as a combination of two losses. We use the Wasserstein distance $\mathcal{L}_\mathrm{W}$ to directly optimize the Gaussian parameters, while the Jensen-Shannon divergence $\mathcal{L}_\mathrm{JS}$, applied on the heatmap values, is used to regularize the implicit occupancy function $O_{\mathbf{p}}$. Our loss is then
\begin{equation}
    \mathcal{L}=\mathcal{L}_\mathrm{W}+\lambda.\mathcal{L}_\mathrm{JS}\mbox{,}
\end{equation}
with a scalar factor $\lambda$ to balance the two losses.


In details, considering predicted and groundtruth Gaussian distributions $\mathcal{G}_\mathrm{pd}$ and $\mathcal{G}_\mathrm{gt}$ characterized by means and covariances $(\mu_\mathrm{pd},\Sigma_\mathrm{pd})$ and $(\mu_\mathrm{gt},\Sigma_\mathrm{gt})$, the Wasserstein distance term is given by:
\begin{equation}
\begin{split}
    \mathcal{L}_\mathrm{W}=&\mbox{ }d_W(\mathcal{G}_\mathrm{pd},\mathcal{G}_\mathrm{gt})\\
    =&\left|\left|\mu_\mathrm{pd}-\mu_\mathrm{gt}\right|\right|_2^2+\mathrm{tr}\left(\Sigma_\mathrm{pd}+\Sigma_\mathrm{gt}-2\left(\Sigma_\mathrm{gt}^{\frac{1}{2}}\Sigma_\mathrm{pd}\Sigma_\mathrm{gt}^{\frac{1}{2}}\right)^{\frac{1}{2}}\right)\mbox{.}
\end{split}
\end{equation}
Such closed-form expression avoids relying on handcrafted relative weights to balance the contributions of mean, variance and covariance terms.

The Jensen-Shannon divergence term is defined as:
\begin{equation}
    \mathcal{L}_\mathrm{JS}=\frac{1}{2}D_\mathrm{KL}(\mathcal{G}_\mathrm{pd}\|\mathcal{G}_\mathrm{m})+\frac{1}{2}D_\mathrm{KL}(\mathcal{G}_\mathrm{gt}\|\mathcal{G}_\mathrm{m})\mbox{,}
\end{equation}
where $ \mathcal{G}_\mathrm{m}=\frac{1}{2}(\mathcal{G}_\mathrm{pd}+\mathcal{G}_\mathrm{gt}) $
and $D_\mathrm{KL}(\mathcal{D}_{1}\|\mathcal{D}_{2})$ is the Kullback-Leibler divergence between any distributions $\mathcal{D}_{1}$ and $\mathcal{D}_{2}$, given by
\begin{equation}
D_\mathrm{KL}(\mathcal{D}_{1}\|\mathcal{D}_{2})=\sum_{(i,j)\in I}\mathcal{D}_{1}(i,j)\mbox{ }\mathrm{log}\left(\frac{\mathcal{D}_{1}(i,j)}{\mathcal{D}_{2}(i,j)}\right)\mbox{.}
\end{equation}
That term is computed directly from the implicit function values over pixels ($i,j$) of heatmap $I$. The Jensen-Shannon divergence has the advantage of being symmetric, in contrast with Kullback-Leibler, and it has been proven to perform better than the latter in~\cite{DSNT}.

\section{EXPERIMENTS AND DISCUSSION}
\label{sec:xp}

\begin{table*}[t]
    \centering
    \normalsize
    \begin{tabular}{l c c c c c}
        & IoU ($\uparrow$) & Overlap ($\uparrow$) & Dice ($\uparrow$) & RVD ($\downarrow$) & MHD ($\downarrow$) \\
        \hline 

        Rubino \textit{et al.}~\cite{RubinoCB18} & 0.78$\pm$0.06 & 0.93$\pm$0.04 & 0.88$\pm$0.04 & 0.12$\pm$0.08 & 6.88$\pm$5.98 \\
        Zins \textit{et al.}~\cite{ZinsSB22_IJCV} & 0.91$\pm$0.12 & 0.96$\pm$0.07 & 0.95$\pm$0.08 & 0.02$\pm$0.04 & 3.93$\pm$5.63 \\
        \hline
                
        Ours ($\mathcal{L}_\mathrm{W}$) & 0.89$\pm$0.05 & \textbf{0.98$\pm$0.02} & 0.94$\pm$0.03 & 0.08$\pm$0.06 & 3.48$\pm$1.80 \\ 
        Ours ($\mathcal{L}_\mathrm{W}+\mathcal{L}_\mathrm{JS}$) & \textbf{0.93$\pm$0.03} & 0.97$\pm$0.02 & \textbf{0.96$\pm$0.01} & \textbf{0.01$\pm$0.01} & \textbf{2.16$\pm$0.66} \\
        Ours ($\mathcal{L}_\mathrm{JS}$) & \textbf{0.93$\pm$0.03} & 0.97$\pm$0.02 & \textbf{0.96$\pm$0.02} & \textbf{0.01$\pm$0.01} & 2.26$\pm$0.77 
    \end{tabular}
    \caption{{\small Experimental validation on AKM~\cite{rathinam2022akm} dataset. Best results are in bold.}}
    \label{tab:res_akm}
\end{table*}

\begin{table*}[t]
    \centering
    \normalsize
    \begin{tabular}{l c c c c c}
        & IoU ($\uparrow$) & Overlap ($\uparrow$) & Dice ($\uparrow$) & RVD ($\downarrow$) & MHD ($\downarrow$) \\
        \hline
        Rubino \textit{et al.}~\cite{RubinoCB18} & 0.70$\pm$0.13 & 0.91$\pm$0.10 & 0.82$\pm$0.13 & 0.17$\pm$0.17 & 88.35$\pm$119.49 \\
        Zins \textit{et al.}~\cite{ZinsSB22_IJCV} & 0.85$\pm$0.18 & \textbf{0.95$\pm$0.12} & 0.91$\pm$0.15 & 0.07$\pm$0.15 & 54.38$\pm$116.38 \\
        \hline
        Ours ($\mathcal{L}_\mathrm{W}$) & 0.82$\pm$0.09 & \textbf{0.95$\pm$0.05} & 0.90$\pm$0.06 & 0.08$\pm$0.13 & 9.59$\pm$7.18 \\
        Ours ($\mathcal{L}_\mathrm{W}+\mathcal{L}_\mathrm{JS}$) & 0.86$\pm$0.08 & \textbf{0.95$\pm$0.05} & 0.92$\pm$0.05 & 0.03$\pm$0.06 & 7.85$\pm$6.86 \\
        Ours ($\mathcal{L}_\mathrm{JS}$) & \textbf{0.87$\pm$0.07} & \textbf{0.95$\pm$0.04} & \textbf{0.93$\pm$0.05} & \textbf{0.01$\pm$0.05} & \textbf{7.58$\pm$6.89}
    \end{tabular}
    \caption{{\small Experimental validation on SPEED+~\cite{SPEED+} dataset. Best results are in bold.}}
    \label{tab:res_sp}
\end{table*}

\textbf{Datasets Extension with 3D-Aware Gaussian Occupancy Labels } We have extended three public spacecraft pose estimation datasets with Gaussian occupancy labels. SPEED~\cite{SPEED} and SPEED+~\cite{SPEED+} are the standard benchmarks for spacecraft pose estimation methods, while AKM~\cite{Rathinam2022} is a recently released dataset featuring texture-less and symmetrical space objects~\cite{pauly2023survey}. No CAD data is available in the first two datasets, but our method requires only the coarse dimensions of the considered spacecraft (TANGO dimensions: 80$\times$75$\times$32cm~\cite{TANGO-spec}).




\vspace{0.1cm}
\textbf{Object Localization Metrics } Our metrics are the Intersection-over-Union score (denoted by IoU), the Overlap score (Overlap), the Dice-Sorensen coefficient (Dice), the Relative Volume Difference (RVD) and the Modified Hausdorff Distance (MHD)~\cite{MHD}. Denoting $P$ and $G$ the sets of pixels inside predicted and groundtruth ellipses, the metrics are defined as
\begin{equation}
    \mathrm{IoU}=\frac{\mathrm{area}(P\cap G)}{\mathrm{area}(P\cup G)}\mbox{,}
\end{equation}
\begin{equation}
    \mathrm{Overlap}=\frac{\mathrm{area}(P\cap G)}{\mathrm{min}(\mathrm{area}(P),\mathrm{area}(G))}\mbox{,}
\end{equation}
\begin{equation}
     \mathrm{Dice}=2\frac{\mathrm{area}(P\cap G)}{\mathrm{area}(P)+\mathrm{area}(G)}\mbox{,}
\end{equation}
\begin{equation}
     \mathrm{RVD}=\frac{|\mathrm{area}(P)-\mathrm{area}(G)|}{\mathrm{area}(G)}\mbox{.}
\end{equation}
While these four metrics characterize in different ways the discrepancy between regions $P$ and $G$ bounded by predicted and groundtruth ellipses, MHD~\cite{MHD} direclty measures the discrepancy between the (discretized) ellipses $\partial P$ and $\partial G$:
\begin{equation}
    \mathrm{MHD}=\mathrm{max}\left(\mathrm{mhd}(\partial P,\partial G),\mathrm{mhd}(\partial G,\partial P)\right)\mbox{,}
\end{equation}
where $\mathrm{mhd}(.)$ is the relative modified Hausdorff distance:
\begin{equation}
    \mathrm{mhd}(\partial P,\partial G)=\frac{1}{|\partial P|}\sum_{p\in\partial P}\min_{g\in\partial G}(\|p-g\|_2)
\end{equation}
in which $p$ and $g$ denote points on the respective discretized ellipses (total number of points for ellipse $\partial P$: $|\partial P|$).

Considering these five metrics allows for a fairly extensive comparison between the performance of different ellipse prediction methods.

\vspace{0.1cm}
\textbf{Baselines } Given a 3D object modelled by an ellipsoid, our method aims at regressing its elliptic projections in the pictures. We provide a qualitative and quantitative comparison with two other representative 3D-aware ellipse regression methods. It is important noting that most 2D ellipse regression methods~\cite{WangDR0XX19,LinLSSS19,Li19,RenHGS15,PanFWZR21,DongRPI21,Wang_2022_ACCV} are, by contrast, to detect naturally occurring 2D elliptic shapes in images.  The first baseline, from Zins \textit{et al.}~\cite{ZinsSB22_IJCV}, is a 2-stage approach consisting in localizing the object using an object detector (Faster R-CNN~\cite{RenHGS15} with ResNet50~\cite{7780459} backbone in the official implementation), then regressing the geometric ellipse parameters from the cropped image. This method is an improved version of their previous work~\cite{ZinsSB20}. The second approach, used in most ellipsoid-based pose estimation problems (\textit{e.g.}, ~\cite{RubinoCB18}), consists in fitting an axis-aligned ellipse within the detected bounding box (same detection model in the experiments). 

\vspace{0.1cm}
\textbf{Object Localization Results } Tables \ref{tab:res_akm} and \ref{tab:res_sp} show a comparison between our model optimized with Wasserstein loss only, Jensen-Shannon loss only and a combination of both ($\lambda=1$), along with methods from Rubino \textit{et al.}~\cite{RubinoCB18} and Zins \textit{et al.}~\cite{ZinsSB22_IJCV}. It shows that our method, even if relying on a lighter backbone (ResNet32), outperforms its competitors on AKM and SPEED+ datasets for each of the five metrics. For our model on AKM dataset, both losses leveraging Jensen-Shannon divergence achieve the same level of accuracy, suggesting that the most important optimization factor is the implicit function regularization. On SPEED+, the model trained only with Jensen-Shannon divergence obtained the best performance because the Gaussian implicit function is slightly truncated outside image boundaries, thus misleading the parameters extraction.

\begin{figure}[!t]
    \centering
    \includegraphics[width=0.45\columnwidth ,trim=0mm 0mm 0mm 0mm, clip]{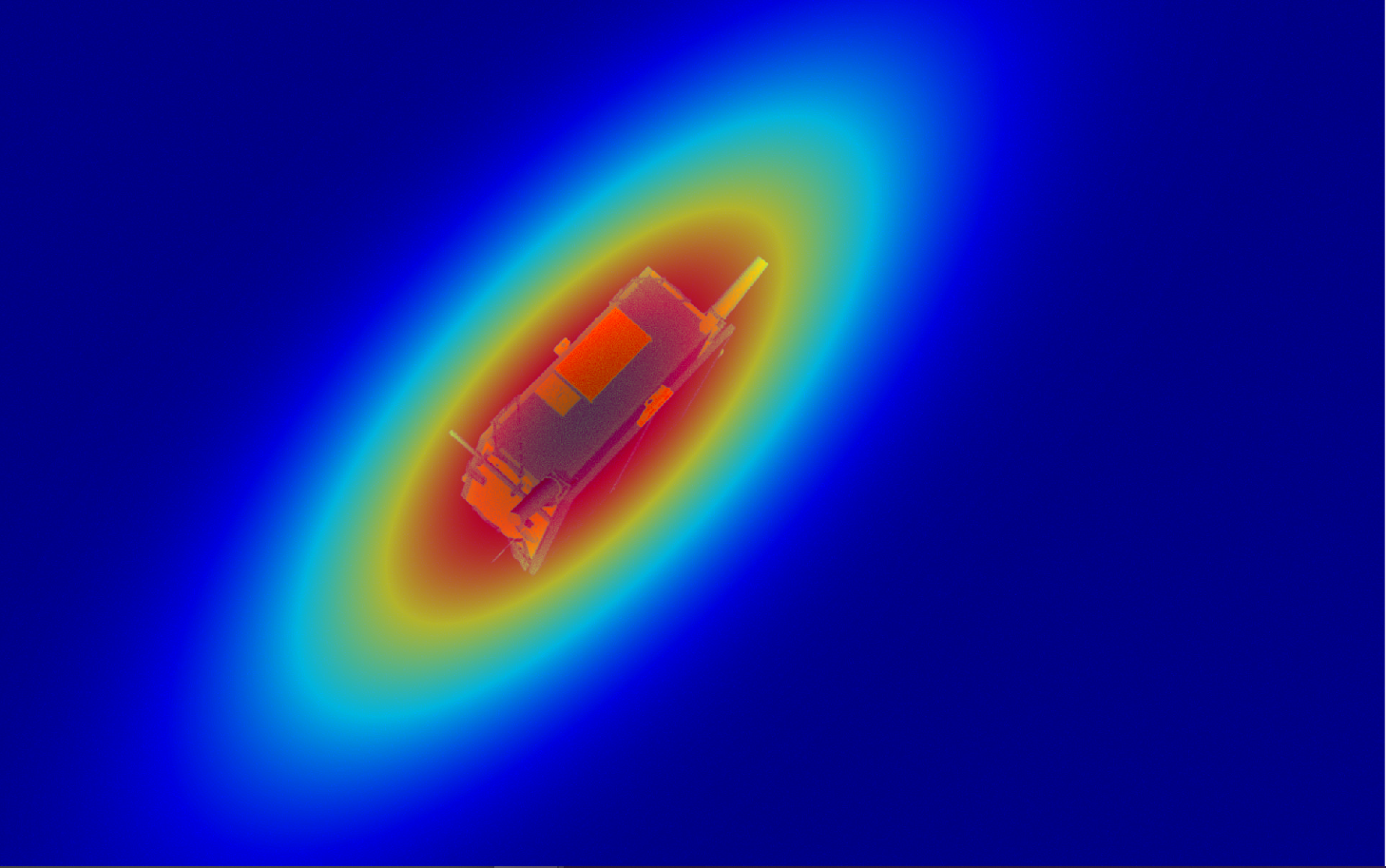}
    \includegraphics[width=.45\columnwidth,trim=0mm 0mm 0mm 0mm, clip]{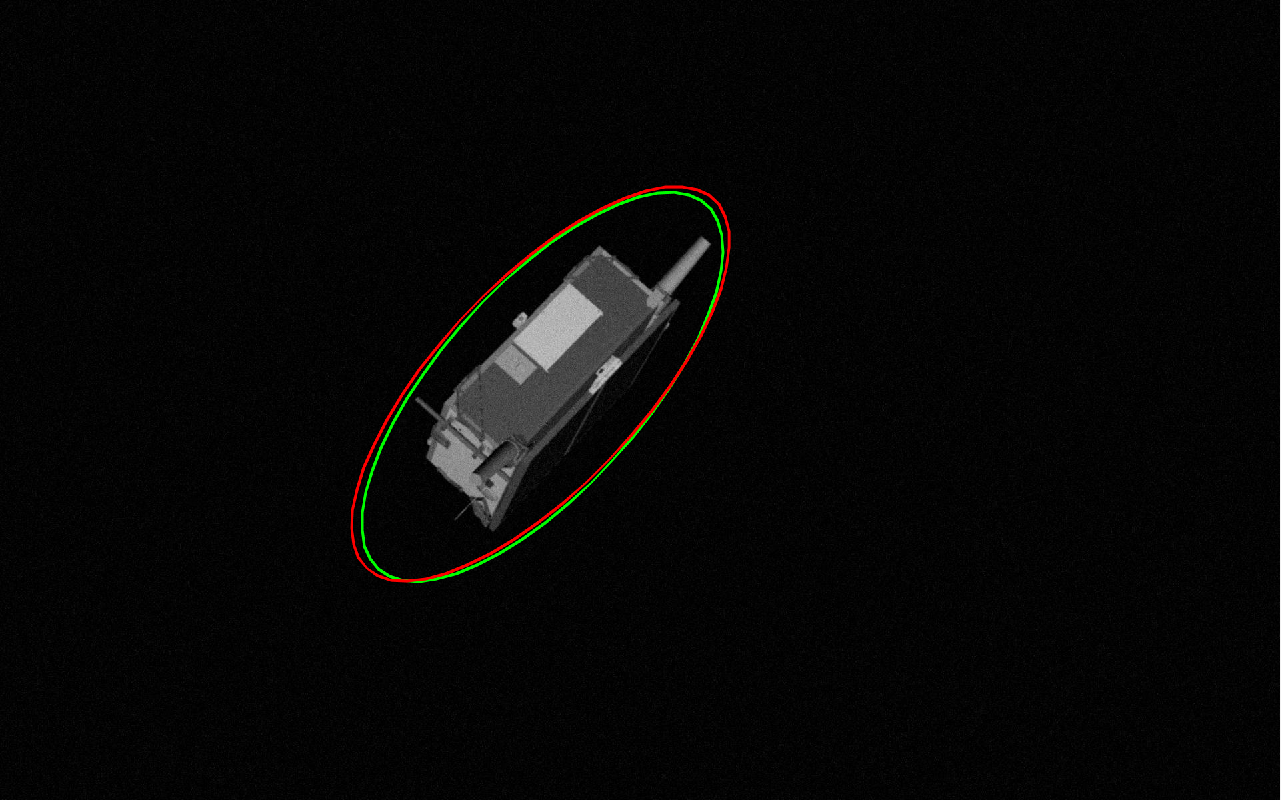}\\
    \vspace{0.1cm}
    \includegraphics[width=.45\columnwidth,trim=0mm 0mm 0mm 0mm, clip]{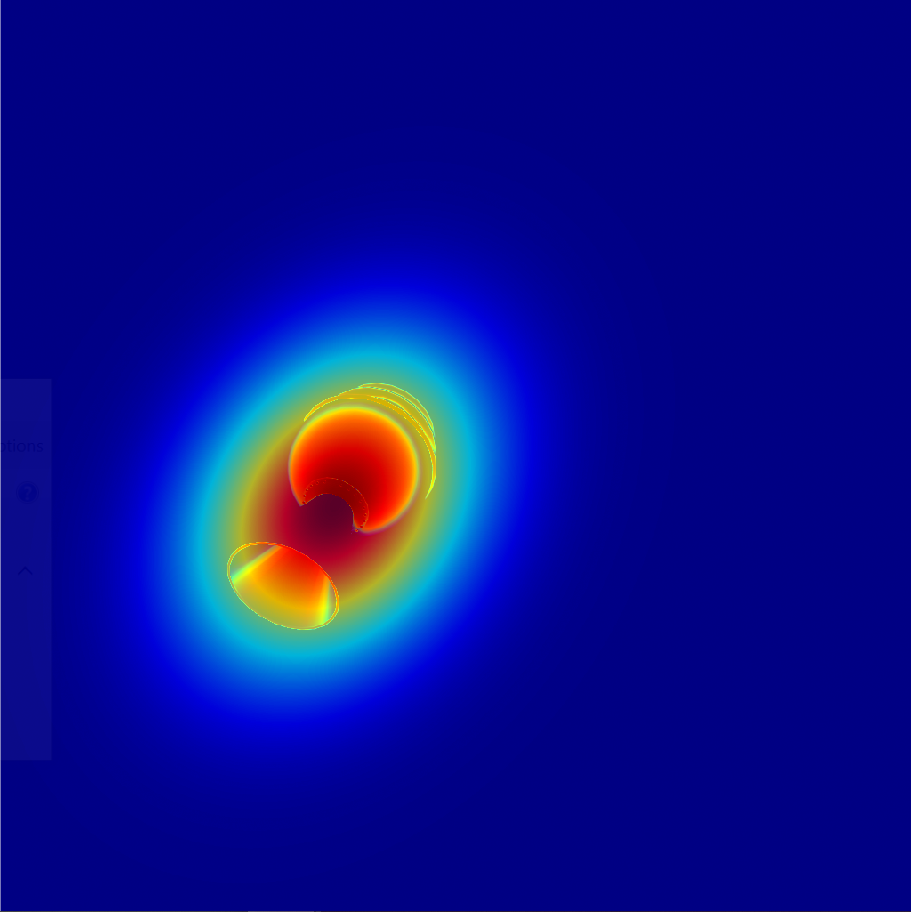}
    \includegraphics[width=.45\columnwidth,trim=0mm 0mm 0mm 0mm, clip]{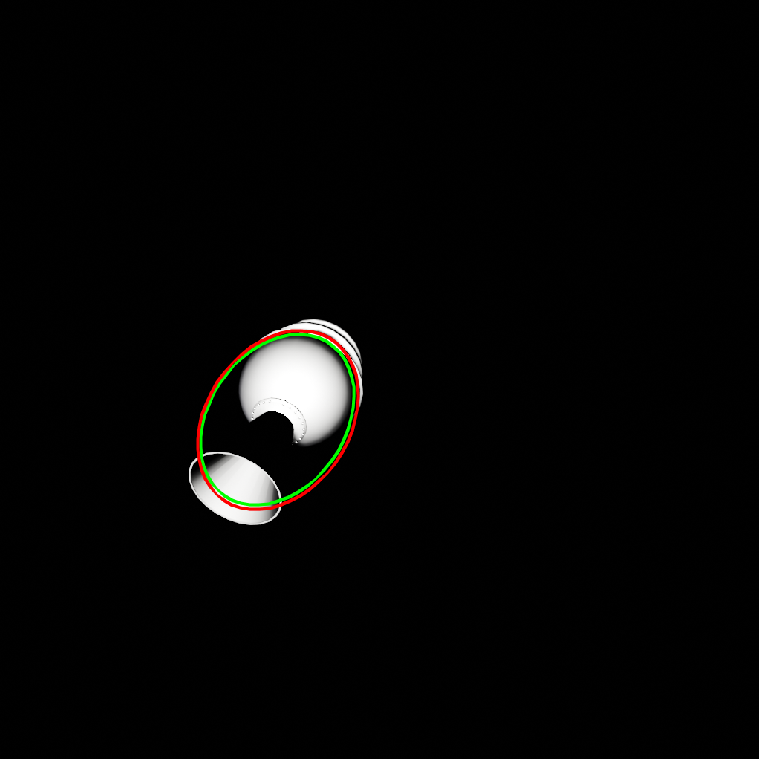}
    \caption{Qualitative results on SPEED+ (top) and AKM (bottom) datasets. Implicit occupancy functions from regressed parameters overlaid as heatmaps (left), with extracted (red) and groundtruth (green) ellipses (right).}
    \label{fig:qual-res}
\end{figure}

In addition, our 1-stage approach has the advantage of performing localization and parameter extraction simultaneously in a fully differentiable manner. Moreover, the parameters extraction is performed by the novel E-DSNT non-trainable layer, hence resulting in a lighter model. Finally, it avoids regressing a discontinuous angular parameter (ellipse orientation), unlike~\cite{ZinsSB20,ZinsSB22_IJCV} and all other ellipse detection methods. Qualitative results, presented in Fig. \ref{fig:qual-res}, show the implicit occupancy function values, groundtruth and predicted ellipses.

\begin{figure}
    \centering
    \includegraphics[trim=0mm 0mm 0mm 0mm, clip, width=\linewidth]{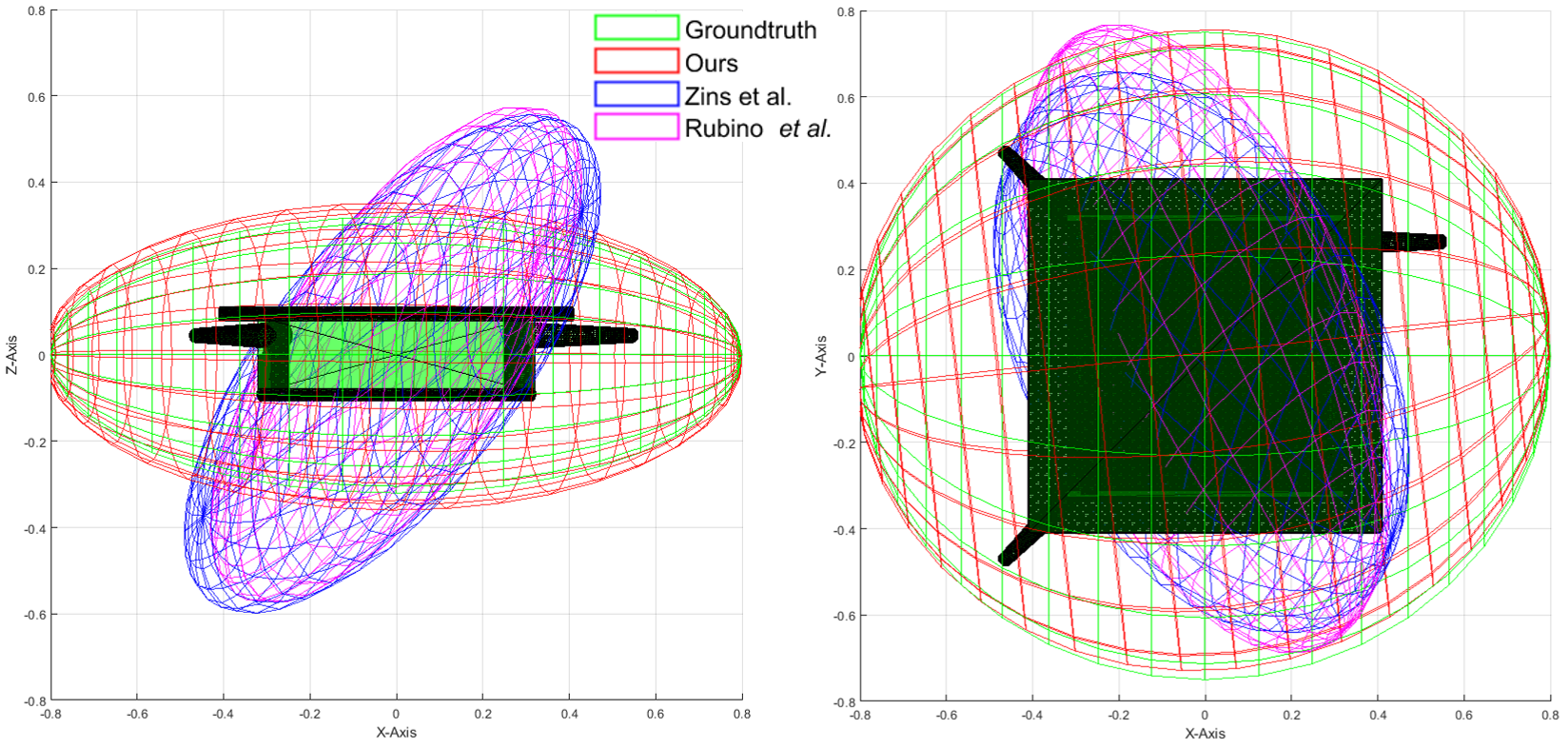}
    \caption{Illustration of TANGO spacecraft (SPEED+ dataset) with groundtruth (in green) and reconstructed ellipsoids from our method (red), Zins \textit{et al.}~\cite{ZinsSB20} (blue) and Rubino \textit{et al.}~\cite{RubinoCB18} (magenta). The reconstruction method is from \cite{RubinoCB18}. The ellipsoid reconstructed from our predictions is closer to the groundtruth ellipsoid than those obtained from other methods.}
    \label{fig:recons}
\end{figure}

\vspace{0.1cm}
\textbf{3D Reconstruction} To assess the 3D-awareness of our predictions, we use regressed ellipse parameters as inputs to a 3D ellipsoid reconstruction method based on triangulation of 2D ellipses~\cite{RubinoCB18}. We randomly selected 100 images from the SPEED+ dataset and reconstructed the ellipsoid based on the ellipses regressed from the different methods. Figure \ref{fig:recons} shows that the ellipsoid reconstructed from our predictions (in red) is closer to the groundtruth ellipsoid (green) than those obtained from other methods (blue, magenta). Quantitatively, an evaluation conducted over 200 subsets of 50 random images is provided in Table \ref{tab:recons_res}. It shows that our method allows for the most accurate reconstruction, demonstrating the 3D awareness of our regressed ellipses, which is of particular importance for a possibly following 6 Degrees-of-Freedom (6DoF) object pose estimation task~\cite{garcia2021lspnet,Gaudillière2023}.

\begin{table}[t]
    \centering
    \begin{tabular}{l c c c}
        & Position & Orientation & Size \\
        \hline
        Rubino \textit{et al.}~\cite{RubinoCB18} & 5.9cm & 10.2$\degree$ & 11.0cm \\
        Zins \textit{et al.}~\cite{ZinsSB22_IJCV} & 2.1cm & 4.1$\degree$ & 8.0cm \\
        Ours ($\mathcal{L}_\mathrm{JS}$) & \textbf{1.0cm} & \textbf{0.1$\degree$} & \textbf{2.4cm}
    \end{tabular}
    \caption{3D reconstruction errors on SPEED+~\cite{SPEED+} dataset, using ellipse triangulation method~\cite{RubinoCB18}. Best results are in bold.}
    \label{tab:recons_res}
\end{table}



\section{CONCLUSION}
\label{sec:concl}

In this paper, we presented a fully differentiable 3D-aware object localization method based on Gaussian implicit occupancy function and ellipse labels. We explained how to readily generate consistent Gaussian occupancy labels to extend already existing pose datasets without requiring any CAD model of the object. We also release the labels for three public spacecraft pose estimation datasets. Future work will focus on integrating that 2D localization model into an end-to-end 6DoF spacecraft pose estimation pipeline, and evaluate on real datasets~\cite{SPEED+,rathinam_arunkumar_2022_6599762}.










\bibliographystyle{./IEEEtran} 
\bibliography{./IEEEexample}

\end{document}